\documentclass[11pt,a4paper]{article}
\usepackage[hyperref]{eacl2021}
\usepackage{times}
\usepackage{placeins}
\usepackage{latexsym}

\usepackage{times}
\usepackage{url}
\usepackage{latexsym}
\usepackage{amsfonts}
\usepackage{amsmath}
\usepackage{multirow}
\usepackage{graphicx}
\usepackage{fixltx2e}
\usepackage{caption}
\usepackage{subcaption}
\usepackage{xspace}
\usepackage{color}
\usepackage{xcolor}
\usepackage{soul}
\usepackage{cleveref}
\usepackage{afterpage}
\usepackage{wrapfig}
\usepackage{lipsum}
\usepackage{booktabs}
\usepackage{enumerate}
\usepackage[export]{adjustbox}
\usepackage{natbib}
\usepackage{enumitem}
\usepackage{float}
\restylefloat{table}

\usepackage{enumitem}

\crefname{section}{§}{§§}
\Crefname{section}{§}{§§}

\usepackage[compact]{titlesec}
\titlespacing{\section}{0pt}{1ex}{0ex}
\titlespacing{\subsection}{0pt}{0.5ex}{0ex}
\titlespacing{\subsubsection}{0pt}{0.5ex}{0ex}

\usepackage{microtype}

\aclfinalcopy 

\title{Adv-OLM: Generating Textual Adversaries via OLM}

\author{Vijit Malik\qquad
  Ashwani Bhat\qquad
  \large{\textbf{Ashutosh Modi}} \\
{Indian Institute of Technology Kanpur (IIT Kanpur)} \\
  {\tt \{vijitvm, bashwani\}@iitk.ac.in}  \\
  {\tt ashutoshm@cse.iitk.ac.in}  \\
}

\date{}

\begin{document}
\maketitle

\begin{abstract}
Deep learning models are susceptible to adversarial examples that have imperceptible perturbations in the original input, resulting in adversarial attacks against these models. Analysis of these attacks on the state of the art transformers in NLP can help improve the robustness of these models against such adversarial inputs. In this paper, we present \textbf{Adv-OLM}, a black-box attack method that adapts the idea of \textbf{O}cclusion and \textbf{L}anguage \textbf{M}odels (OLM) to the current state of the art attack methods. OLM is used to rank words of a sentence, which are later substituted using word replacement strategies. We experimentally show that our approach outperforms other attack methods for several text classification tasks.
\end{abstract}

\section{Introduction}
\noindent In recent times, deep learning models have become pervasive across different domains.
Many of the recent deep models have shown SOTA performance on a variety of NLP tasks \citep{wang2018glue}. 
Consequently, deep models are being deployed in a variety of production systems for real-life applications. 
Hence, it becomes imperative to ensure the reliability and robustness of such models as it might pose a threat to security.

Recent studies have pointed out the vulnerability of deep models to adversarial attacks \citep{goodfellow2014explaining}. Adversarial attack comprises generating adversarial samples by performing small perturbations to the original input, making them imperceptible to humans while fooling the deep learning models to give incorrect predictions. 

Adversarial attack on textual data is much more difficult due to the discrete nature of the text. The basic requirement of imperceptibility of perturbation by human judges is much more challenging in a language data setting. Therefore, the adversarial sample needs to be grammatically correct and semantically sound.
Perturbations at word or character level that are perceptible to human judges have been explored in-depth \citep{ebrahimi2017hotflip,belinkov2017synthetic,jia2017adversarial,gao2018black}. Work on defense against misspellings based attacks \citep{pruthi2019combating} and use of optimization algorithms for attacks like genetic algorithm \citep{alzantot2018generating,wang2019natural} and particle swarm optimization \citep{zang2020word} have also been explored. 
With the rise of pre-trained language models, like BERT \citep{devlin2018bert} and other transformer-based models, generating human imperceptible adversarial examples has become more challenging.  \citet{wallace2019universal},  \citet{jin2019bert}, and \citet{pruthi2019combating} have explored these models from different perspectives.

Adversarial examples can be generated using black-box, where no knowledge about the model is accessible, and white-box, where information about the technical details of models are known.
Generation of textual adversarial samples in a black-box setting consists of two steps 1) Finding words to replace in a sample (Word Ranking) 2) Replacing the chosen word (Word Replacement). Word Ranking is necessary to ensure that the word that contributes the most to the output prediction is considered as the candidate for replacement in the next step. Other constraints like generating semantically similar adversarial samples, human imperceptibility, and minimal perturbation percentage are also considered. Previous work has obtained word ranking by performing deletion of words 
(e.g., BAE-R \citep{garg2020bae}, TextFooler \cite{jin2019bert}),  and replacement of words with \textit{[UNK]} token (e.g., BERT-Attack \citep{li2020bert}) and then ranking the words based on the output logits difference. 

Recently in the model explainability domain, the method of Occlusion and Language Models (OLM) \citep{harbecke2020considering} has been proposed, the authors argue that the data likelihood of the samples obtained after either deleting the token or replacement with \textit{[UNK]} token is very low, which makes these methods unsuitable for determining relevance of the word towards the output probability. The authors propose the use of language models for calculating the relevance of the words in a sentence.
Taking inspiration from OLM, we propose \textbf{Adv-OLM}, a black box attack method, that adapts the idea of OLM (as the Work Ranking Strategy) to find the relevant words to replace.
We empirically show that OLM provides a better set of ranked words compared to the existing word ranking strategies for the generation of adversarial examples.

\noindent We summarize our contributions as follows:
\begin{itemize}[noitemsep,topsep=-2pt]
    \item We propose a new method Adv-OLM, to rank words for generating adversarial examples.
    \item We empirically show that Adv-OLM has a higher success rate and lower perturbation percentage than previous attacking methods.
\end{itemize}

The implementation for the proposed approach is made available at the GitHub repository: \url{https://github.com/vijit-m/Adv-OLM}.

\section{Problem Formulation}
\noindent We are given a corpus consisting of $n$ input samples, $\mathbb{X} = \{x_1,\ldots , x_n\}$ with corresponding labels $\mathbb{Y} = \{y_1, \ldots , y_n\}$ and a trained classification model $f (f : \mathbb{X} \rightarrow \mathbb{Y})$ that maps an input samples to its correct label. We assume a black-box setting where the attacker can only query the classifier for output label probabilities for the given input. For an input sample $x \in \mathbb{X}$, the task is to construct an adversarial sample $x^{\prime}$ such that, $f(x) = y, \quad f(x^{\prime}) = y^{\prime}$ with $\quad y \neq y^{\prime}$, and $Similarity(x^{\prime},x) \geq \epsilon$. 
Here, $Similarity: \mathbb{X}\times\mathbb{X}\rightarrow(0,1)$ can be both the semantic and syntactic similarity function, and $\epsilon$ is the minimum similarity threshold. Ideally, the amount of perturbation should be minimized. 
The first step is to rank the words of the sample $x$.
Based on the ranking, starting from the most important word, the word is replaced by some candidate word that keeps the perturbed sample $x^{\prime}$ semantically similar and grammatically sound but changes the output prediction.


\begin{figure*}[!ht]
\centering
\begin{subfigure}{\textwidth}
  \includegraphics[width=1\linewidth]{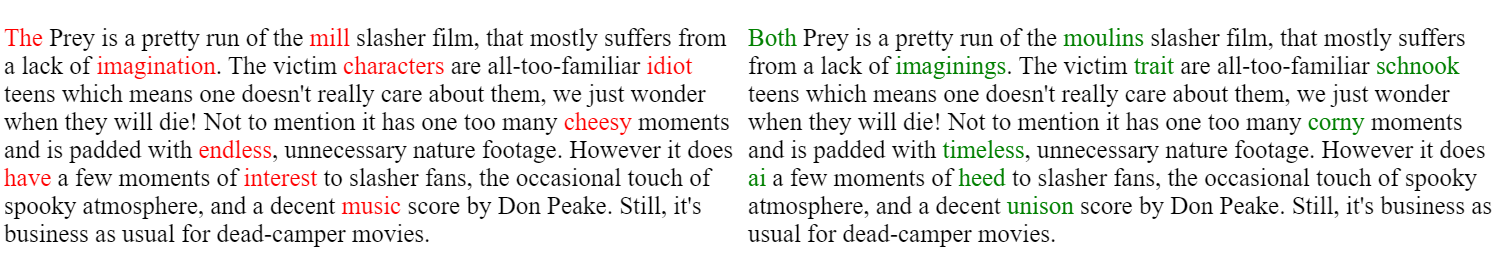}
  \caption{TextFooler Attack on fine-tuned BERT on IMDB data sample. $[\textcolor{red}{Negative (100\%)} \rightarrow \textcolor{green}{Positive (59\%)}]$
  }
  \label{fig:textfooler_imdb}
\end{subfigure}

\begin{subfigure}{\textwidth}
  \includegraphics[width=1\linewidth]{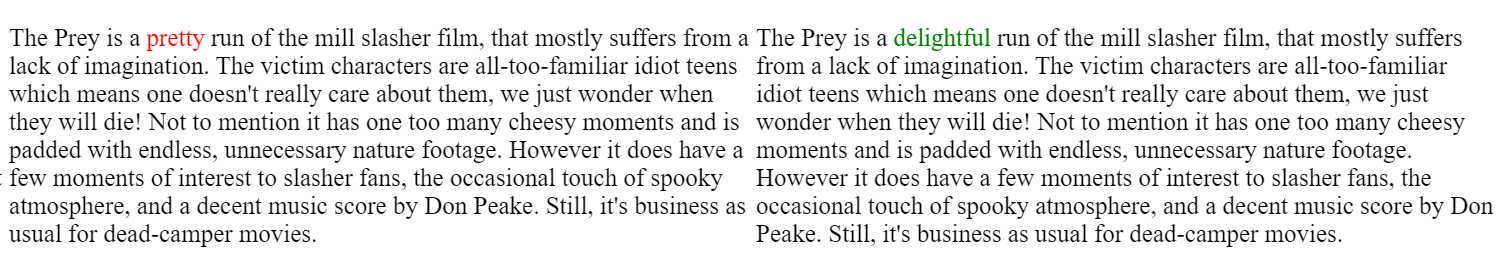}
  \caption{Adv-OLM Attack on fine-tuned BERT on IMDB data sample.
    $[\textcolor{red}{Negative (100\%)} \rightarrow \textcolor{green}{Positive( 95\%)}]$
    }
    \label{fig:textfooler_olm_imdb}
\end{subfigure}
\caption{Qualitative Examples of TextFooler and Adv-OLM on BERT classifier (\textcolor{red}{Red} words are replaced by \textcolor{green}{Green} words while changing the output prediction probability.)}
\label{fig:textfooler_olm_example}
\end{figure*}
\section{Methodology}
\noindent Adv-OLM uses the idea of Occlusion and Language Models to perform Word Ranking using both OLM and OLM-S methods. 
OLM uses a language model to sample some candidate instances for a word and then replaces the word. Let $x_{i}$ be a word of the input $x$ and $x_{\backslash i}$ be the incomplete input without this word. Then the OLM relevance score $r$ given the prediction function $f$ and label $y$ is (Here $f_{y}$ is the logit value corresponding to the label $y$.)
\setlength{\belowdisplayskip}{4pt} \setlength{\belowdisplayshortskip}{4pt}
\setlength{\abovedisplayskip}{4pt} \setlength{\abovedisplayshortskip}{4pt}
\begin{equation} \label{eq:OLMrel}
r_{f,y}(x_{i}) = f_{y}(x_{i}) - f_{y}(x_{\backslash i})
\end{equation}
Here, $f_{y}(x_{\backslash i})$ is not accurately defined and needs to be approximated since $x_{\backslash i}$ is the  incomplete input. A language model $p_{LM}$ generates input by predicting the masked word as $\hat{x_{i}}$ that is as natural as possible for the model and thus approximates to:
\begin{equation} \label{eq:OLMlm}
f_{y}(x_{\backslash i}) \approx \sum_{\hat{x_{i}}} p_{LM}(\hat{x_{i}}|x_{\backslash i})f_{y}(x_{\backslash i} \cup \hat{x_{i}}) 
\end{equation}
where, $f_{y}(x_{\backslash i} \cup  \hat{x_{i}}) $ is the prediction of the classification model after the language model's prediction $\hat{x_{i}}$ is added to the incomplete input $x_{\backslash i}$.  

The other method OLM-S calculates the sensitivity of a position in the text and has nothing to do with the word present at that position in the original input. The sensitivity score of OLM-S is calculated
\begin{equation*} \label{eq:OLMSsensi}
\resizebox{.95\hsize}{!}{$
s_{f,y}(x_{i})=\sqrt{\sum_{\hat{x_{i}}} p_{LM}(\hat{x_{i}}|x_{\backslash i})(f_{y}(x_{\backslash i}\cup \hat{x_{i}})-\mu)^{2}} $}
\end{equation*}
where $\mu$ is the mean value from Equation \ref{eq:OLMlm}.
The sensitivity score $s_{f,y}(x_{i})$ is used for word ranking in OLM-S.

After performing the Word Ranking step using the relevance scores generated by OLM and OLM-S, the next step is to replace highly scored words with semantically similar words that form grammatically correct sentences (Word Replacement) such that the output prediction changes. Word replacement strategy is kept similar to existing methods. TextFooler uses Synonym Extraction, POS checking and semantic similarity checking whereas BAE-R uses a Language Model for word replacement. (details in Appendix \ref{section:wordreplacement}).

\begin{table}[!h]
\centering
\resizebox{0.8\columnwidth}{!}{
\begin{tabular}{|l|l|l|l|l|}
\hline
\multicolumn{1}{|c|}{\textbf{Dataset}}              & \multicolumn{1}{c|}{\textbf{Classes}} & \multicolumn{1}{c|}{\textbf{Train}} & \multicolumn{1}{c|}{\textbf{Test}} & \multicolumn{1}{c|}{\textbf{\begin{tabular}[c]{@{}c@{}}Avg. Length
\end{tabular}}} \\ \hline \hline
\multicolumn{5}{|c|}{\textbf{Classification}} \\ \hline 
IMDB & 2 & 25K & 500 & 245.12                   
\\ \hline
Yelp & 2 & 560K & 500 & 132.33                  
\\ \hline
\begin{tabular}[c]{@{}l@{}}AG's News\end{tabular} & 4 & 120K & 500 & 40.41                          \\ \hline
\multicolumn{5}{|c|}{\textbf{Natural Language Inference}}       
\\ \hline
MNLI & 3 & 433K & 500 & 29.72                   
\\ \hline
\end{tabular}
}
\caption{Statistic of Datasets. Avg. Length is the average number of words in the test set.}
\label{table:dataset}
\end{table}


\begin{table*}[t]\small
\renewcommand{\arraystretch}{1.5}
\centering

\setlength\tabcolsep{3.5pt}
\resizebox{2\columnwidth}{!}{
\begin{tabular}{|c|c|c|c|ccc|c|ccc|}
\hline
\multirow{2}{*}{\textbf{Dataset}} & \multirow{2}{*}{\textbf{Method}} & \multirow{2}{*}{\textbf{Word Ranking}} & \multicolumn{4}{c|}{\textbf{BERT}} & \multicolumn{4}{c|}{\textbf{ALBERT}}              \\ \cline{4-11} &  &  & 
\textbf{Original Acc.}  & \multicolumn{1}{c|}{\textbf{Attacked Acc.}} & \multicolumn{1}{c|}{\textbf{Success Rate}} & \multicolumn{1}{c|}{\textbf{Perturbed \%}} & \textbf{Original Acc.}  & \multicolumn{1}{c|}{\textbf{Attacked Acc.}} & \multicolumn{1}{c|}{\textbf{Success Rate}} & \multicolumn{1}{c|}{\textbf{Perturbed \%}} \\ \hline \hline
\multirow{7}{*}{\textbf{\begin{tabular}[c]{@{}c@{}}AG's\\ News\end{tabular}}} & 

\multirow{3}{*}{BAE-R} & OLM  & \multirow{7}{*}{93.8\%} & \multicolumn{1}{c|}{\textbf{78\%}}  & \multicolumn{1}{c|}{\textbf{16.84\%}}  & \multicolumn{1}{c|}{6.72\%}   & \multirow{7}{*}{94.4\%} & \multicolumn{1}{c|}{\textbf{79.2\%}}  & \multicolumn{1}{c|}{\textbf{16.1\%}} & \multicolumn{1}{c|}{8.37\%} \\
 &   & OLM-S   &    & \multicolumn{1}{c|}{79\%}  & \multicolumn{1}{c|}{15.78\%}  & \multicolumn{1}{c|}{\textbf{6.35\%}}  &  & \multicolumn{1}{c|}{82.4\%}  & \multicolumn{1}{c|}{12.71\%} & \multicolumn{1}{c|}{\textbf{6.9\%}} \\
 &  & Original (delete) &    & \multicolumn{1}{c|}{78.8\%}   & \multicolumn{1}{c|}{15.99\%}     & \multicolumn{1}{c|}{6.42\%}  &  & \multicolumn{1}{c|}{79.4\%}  & \multicolumn{1}{c|}{15.89\%}  & \multicolumn{1}{c|}{7.67\%}  \\ 
 \cline{2-3} \cline{5-7} \cline{9-11} 
 
 & \multirow{3}{*}{TextFooler} & OLM  &  & \multicolumn{1}{c|}{\textbf{19.2\%}}    &    \multicolumn{1}{c|}{\textbf{79.53\%}}     & \multicolumn{1}{c|}{23.52\%}  &    & \multicolumn{1}{c|}{20.4\%} & \multicolumn{1}{c|}{78.39\%}  & \multicolumn{1}{c|}{21.24\%} \\   
 &  & OLM-S  &  & \multicolumn{1}{c|}{21.4\%}  & \multicolumn{1}{c|}{77.19\%} & \multicolumn{1}{c|}{\textbf{20.96\%}}  &   &\multicolumn{1}{c|}{\textbf{20.2\%}}  & \multicolumn{1}{c|}{\textbf{78.6\%}} & \multicolumn{1}{c|}{\textbf{20.18\%}} \\ 
 &  & Original (delete)  &   & \multicolumn{1}{c|}{21.4\%}  & \multicolumn{1}{c|}{77.19\%} & \multicolumn{1}{c|}{23.19\%}   & & \multicolumn{1}{c|}{22.0\%} & \multicolumn{1}{c|}{76.69\%}  & \multicolumn{1}{c|}{21.15\%} \\ 
 \cline{2-3} \cline{5-7} \cline{9-11}
 & PWWS  & - &  & \multicolumn{1}{c|}{44.8\%}  & \multicolumn{1}{c|}{52.24\%} & \multicolumn{1}{c|}{16.21\%}    &    & \multicolumn{1}{c|}{36.8\%}  & \multicolumn{1}{c|}{61.02\%}  & \multicolumn{1}{c|}{14.8\%}  \\ \hline
 
\multirow{7}{*}{\textbf{Yelp}} & \multirow{3}{*}{BAE-R} 
& OLM  & \multirow{7}{*}{97.2\%} & \multicolumn{1}{c|}{\textbf{38.4\%}} & \multicolumn{1}{c|}{\textbf{60.49\%}} & \multicolumn{1}{c|}{\textbf{6.45\%}} & \multirow{7}{*}{97.4\%} & \multicolumn{1}{c|}{41.4\%} & \multicolumn{1}{c|}{57.49\%} & \multicolumn{1}{c|}{7.07\%} \\ 
&  & OLM-S  & & \multicolumn{1}{c|}{55.2\%}  & \multicolumn{1}{c|}{43.21\%}  & \multicolumn{1}{c|}{10.36\%}  &   & \multicolumn{1}{c|}{39.6\%}  & \multicolumn{1}{c|}{59.34\%} & \multicolumn{1}{c|}{7.11\%} \\ 
&  & Original (delete) &  & \multicolumn{1}{c|}{41.6\%}  & \multicolumn{1}{c|}{57.20\%}  & \multicolumn{1}{c|}{7.28\%}   &   & \multicolumn{1}{c|}{35.0\%} & \multicolumn{1}{c|}{64.07\%}  & \multicolumn{1}{c|}{6.49\%} \\ \cline{2-3} \cline{5-7} \cline{9-11} 
& \multirow{3}{*}{TextFooler}      
& OLM &  & \multicolumn{1}{c|}{\textbf{5.2\%}} & \multicolumn{1}{c|}{\textbf{94.65\%}} & \multicolumn{1}{c|}{\textbf{9.10\%}}  &   & \multicolumn{1}{c|}{\textbf{2.6\%}}         & \multicolumn{1}{c|}{\textbf{97.33\%}}  & \multicolumn{1}{c|}{10.17\%}  \\ 
&  & OLM-S   &   & \multicolumn{1}{c|}{7.8\%}   & \multicolumn{1}{c|}{91.98\%}  & \multicolumn{1}{c|}{13.31\%}  &   & \multicolumn{1}{c|}{2.8\%}   & \multicolumn{1}{c|}{97.13\%} & \multicolumn{1}{c|}{10.13\%} \\
&  & Original (delete)   &   & \multicolumn{1}{c|}{6.6\%} & \multicolumn{1}{c|}{93.21\%}   & \multicolumn{1}{c|}{9.95\%}  & & \multicolumn{1}{c|}{3.8\%}  & \multicolumn{1}{c|}{96.1\%}  & \multicolumn{1}{c|}{9.57\%} \\ \cline{2-3} \cline{5-7} \cline{9-11}
& PWWS  & -   &    & \multicolumn{1}{c|}{6.2\%}  & \multicolumn{1}{c|}{93.62\%}  & \multicolumn{1}{c|}{6.9\%}   &   & \multicolumn{1}{c|}{3.8\%}   & \multicolumn{1}{c|}{96.1\%}   & \multicolumn{1}{c|}{6.82\%} \\ \hline
\end{tabular}
}
\caption{Comparison between word ranking strategies on AG's News and Yelp for fine-tuned BERT and ALBERT. Our method Adv-OLM has OLM (or OLM-S) as the word ranking strategy.}
\label{tab:bertandalbert}

\end{table*}


\begin{table*}[h!]\small
\renewcommand{\arraystretch}{1.5} 
\centering
\setlength\tabcolsep{3.5pt}
\resizebox{1.3\columnwidth}{!}{
\begin{tabular}{|c|c|c|c|c|c|c|}
\hline
\textbf{\begin{tabular}[c]{@{}c@{}}Dataset\end{tabular}} & 
\textbf{Method} & 
\textbf{\begin{tabular}[c]{@{}c@{}}Word Ranking\end{tabular}} & \textbf{\begin{tabular}[c]{@{}c@{}}Original Acc.\end{tabular}} &
\textbf{\begin{tabular}[c]{@{}c@{}}Attacked Acc.\end{tabular}} &
\textbf{\begin{tabular}[c]{@{}c@{}}Success Rate\end{tabular}} &
\textbf{\begin{tabular}[c]{@{}c@{}}Perturbed \%\end{tabular}}  
\\ \hline \hline
\multirow{7}{*}{\textbf{\begin{tabular}[c]{@{}c@{}}MNLI\end{tabular}}} &
\multirow{3}{*}{BAE-R} & 
OLM & \multirow{7}{*}{84.6\%} & 20.6\% & 75.65\% & 7.82\%   
\\ 
& & 

OLM-S & & 20.8\% & 75.41\% & 8.22\%
\\ 
& & 

\begin{tabular}[c]{@{}c@{}}Original (delete)\end{tabular} & & 
14.0\% & 83.45\% & 6.4\%                                             \\ \cline{2-3} \cline{5-7} & 

\multirow{3}{*}{TextFooler} & OLM & & 
\textbf{6.6\%} & \textbf{92.2\%} & 8.29\%                           \\
& & 

OLM-S & & 7.0\% & 91.73\% & 8.59\%
\\ 
& & 

\begin{tabular}[c]{@{}c@{}}Original (delete)\end{tabular} & & 
6.8\% & 91.96\% & 6.98\%
\\ \cline{2-3} \cline{5-7} & 

PWWS & - & & 3.2\% & 96.22\% & 6.62\%                               \\ \hline

\end{tabular}
}
\caption{Comparsion between previous methods and Adv-OLM on MNLI fine-tuned BERT. Our method Adv-OLM has OLM (or OLM-S) as the word ranking strategy.}
\label{tab:mnli}
\end{table*}


\begin{table*}[h!]\small
\renewcommand{\arraystretch}{1.5} 
\centering
\setlength\tabcolsep{5pt}
\resizebox{1.4\columnwidth}{!}{
\begin{tabular}{|c|c|c|c|c|c|c|}
\hline
\textbf{Model} &
\textbf{Method} & 
\textbf{
\begin{tabular}[c]{@{}c@{}}Word Ranking\end{tabular}} &
\textbf{Original Acc.}  & 
\textbf{Attacked Acc.} & 
\textbf{Success Rate} & 
\textbf{Perturbed \%} 
\\ \hline \hline

\multirow{7}{*}{\textbf{BERT}} & 
\multirow{3}{*}{BAE-R} & OLM & 
\multirow{7}{*}{92.0\%} & 38.8\% & 57.83\% & 2.79\%
\\ 
& & 

OLM-S & & \textbf{33.2\%} & \textbf{63.91\%} & \textbf{2.33\%} 
\\ 
& & 

Original (delete) & & 42.6\% & 53.70\% & 2.92\%
\\ \cline{2-3} \cline{5-7} & 

\multirow{3}{*}{Textfooler} & 
OLM & & 29.8\% & 67.61\% & 4.52\%
\\
& & 

OLM-S & & \textbf{26.4\%} & \textbf{71.3\%} & \textbf{3.16\%}
\\ 
& & 

Original (delete) &  & 31.8\% & 65.43\% & 5.26\%
\\ \cline{2-3} \cline{5-7} & 

PWWS & \_ & & 28.2\% & 69.35\% & 2.92\% \\ \hline

\multirow{7}{*}{\textbf{ALBERT}} &
\multirow{3}{*}{BAE-R} & 
OLM & \multirow{7}{*}{92.8\%} & 27.4\% & 70.47\% & 3.58\% 
\\ 
& & 

OLM-S & & \textbf{23.2\%}  & \textbf{75.0\%} & \textbf{3.4\%}
\\ 
& & 

Original (delete) & & 26.8\% & 71.12\% & 3.48\%
\\ 
\cline{2-3} \cline{5-7} 
& 

\multirow{3}{*}{Textfooler} & 
OLM & & \textbf{1.4\%} & \textbf{98.49\%} & 6.65\%
\\ 
& & 

OLM-S &  & \textbf{1.4\%} & \textbf{98.49\%} & \textbf{5.39\%} 
\\
& & 

Original (delete) & & 2.4\%  & 97.41\% & 6.57\%
\\ \cline{2-3} \cline{5-7}  &

PWWS & \_ & & 3.8\% & 95.91\% & 3.76\% \\ \hline

\multirow{7}{*}{\textbf{RoBERTa}} & 
\multirow{3}{*}{BAE-R} & 
OLM & \multirow{7}{*}{94.2\%} & 29.4\% & 68.79\% & 4.08\%  
\\
& & 

OLM-S &  & \textbf{28.0\%}  & \textbf{70.28\%} & \textbf{3.57\%}
\\ 
& & 

Original (delete) &  & 29.4\%  & 68.79\% & 3.9\% \\ 
\cline{2-3} \cline{5-7} &

\multirow{3}{*}{Textfooler} & 
OLM  &  & \textbf{0.0\%}    & \textbf{100\%} & 7.62\% 
\\ 
& & 

OLM-S & & 0.2\% & 99.79\% & \textbf{6.43\%} 
\\
& & 

Original (delete) & & 0.2\% & 99.79\% & 6.89\%
\\
& 

PWWS & \_ & & 0.4\% & 99.58\% & 5.38\%  
\\ \hline

\multirow{7}{*}{\textbf{DistilBERT}} & \multirow{3}{*}{BAE-R} & 
OLM & 
\multirow{7}{*}{91.8\%} & 22.6\% & 75.38\% & 3.29\%  
\\ 
& & 

OLM-S & & 21.6\% & 76.47\% & \textbf{2.93\%} 
\\
& & 

Original (delete) & & 21.6\% & 76.47\% & 3.34\% 
\\ \cline{2-3} \cline{5-7} & 

\multirow{3}{*}{Textfooler} & 
OLM & & 0.2\% & 99.78\% & 4.03\% 
\\ 
& & 

OLM-S & & 0.2\% & 99.78\% & \textbf{3.55\%} 
\\ 
& & 

Original (delete) &  & 0.2\%  & 99.78\% & 4.44\% 
\\ \cline{2-3} \cline{5-7} & 

PWWS & \_ & & 0.6\% & 99.35\% & 3.0\%
\\ \hline

\multirow{7}{*}{\textbf{BiLSTM}} & 
\multirow{3}{*}{BAE-R} & 
OLM & \multirow{7}{*}{83.8\%} & 10.8\% & 87.11\% & 2.78\% 
\\ 
& & 

OLM-S & & 10.4\% & 87.59\% & 2.71\% 
\\ 
& & 

Original (delete) & & 8.6\% & 89.74\% & 2.52\% 
\\ \cline{2-3} \cline{5-7} & 

\multirow{3}{*}{Textfooler} & OLM & & 0\% & 100\% & 2.38\% 
\\ 
& & 

OLM-S &  & 0\% & 100\%  & 2.41\%
\\ 
& & 

Original (delete) & & 0\%  & 100\% & 1.95\% 
\\ \cline{2-3} \cline{5-7} & 

PWWS & \_ & & 0\% & 100\% & 1.63\% \\ \hline

\end{tabular}
}
\caption{Comparison between previous methods and Adv-OLM for IMDB dataset across different models. Our method Adv-OLM has OLM (or OLM-S) as the word ranking strategy.}
\label{tab:transformersimdb}
\end{table*}

\section{Experiments}
\label{section:experiments}
\noindent We experiment with different benchmark datasets for text classification and entailment: IMDB, AG News, Yelp Polarity and MNLI (details in Appendix \ref{section:dataset}). The statistics of the final dataset are shown in Table \ref{table:dataset}. Test set was randomly choosen stratified set. For evaluating the effectiveness of our proposed approach, we experiment with SOTA text classifiers i.e. transformer based models like BERT \citep{devlin2018bert}, ALBERT \citep{lan2019albert}, RoBERTa \citep{liu2019roberta} and DistilBERT \citep{sanh2019distilbert}.

We replaced the existing word ranking strategies (i.e. Original (delete)) of previous attack methods: Textfooler \citep{jin2019bert} and BAE-R \citep{garg2020bae} with word rankings generated using OLM and OLM-S while keeping rest of the attack procedure same. The comparison is provided between the attacks generated through original word ranking, and OLM adapted word ranking (including comparison with PWWS attack method \citep{ren2019generating}) in \cref{tab:bertandalbert}, \cref{tab:mnli} and \cref{tab:transformersimdb}. PWWS (Probability Weighted Word Saliency) method considers the word
saliency along with the classification probability. The change in value of the classification probability is used to measure the attack effect of the proposed substitute word, while word saliency shows how well the original word affects the classification. We use the default language model (BERT) employed in the OLM and OLM-S, and 
kept the number of samples generated by the OLM language model as 30 in all the experiments. 

\noindent The following evaluation metrics are used:
\noindent
\begin{itemize}[noitemsep,topsep=-2pt]
    \item \textbf{Attacked Acc.}: Accuracy of the model after attack. Lower the better.
    \item \textbf{Success Rate:} Ratio of number of successful attacks and the total number of attempted attacks\footnote{Note that total number of attempted attacks are not the same as number of input examples i.e., the samples which were originally wrongly classified by the model even before an attack are skipped}. Higher the better.
    \item \textbf{Perturbed Percentage:} Ratio of number of words that were modified by the attack and the total number of words in the input example. Lower the better.
\end{itemize}
We use TextAttack's \citep{morris2020textattack} fine tuned models on these datasets and used it to execute the attacks, including Adv-OLM (Appendix \ref{section:textattack}).

\noindent \textbf{Number of queries in Adv-OLM}:
From equations \ref{eq:OLMlm} and \ref{eq:OLMSsensi}, it is clear that unlike other methods of deletion and \textit{[UNK]} token replacement, which perform only a single query, we need to perform multiple queries. 
\begin{figure}[h!]
    \centering
    \includegraphics[width=0.48\textwidth]{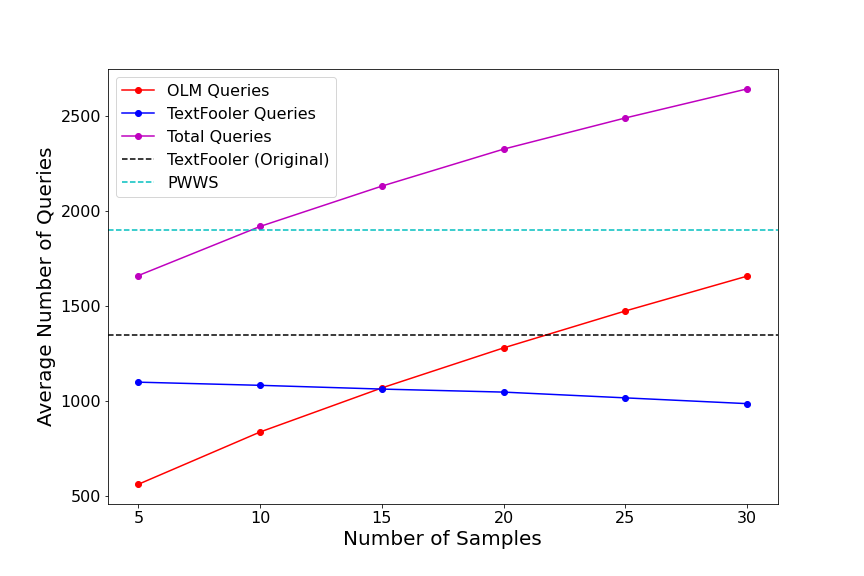}
    \caption{Average Number of Queries vs Number of samples in Adv-OLM attack on BERT on IMDB dataset.}
    \label{fig:queries}
\end{figure}
We set the number of samples generated by the OLM language model to 30 for our experiment. In the worst case, we would have all 30 samples of the token as unique, which will query the model 30 times. However, experimentally it was not the case. To study this, we varied the number of samples and evaluated the OLM ranking step's number of queries. In fig \ref{fig:queries}, we plotted the number of queries for OLM averaged over the input samples against the number of samples. We can see that there is not a significant difference in the total number of samples in our case (OLM + Textfooler queries) when compared with PWWS. 

\section{Results and Analysis}
\noindent Results are shown in Tables \ref{tab:bertandalbert}, \ref{tab:mnli} and \ref{tab:transformersimdb}. Table \ref{tab:bertandalbert} provides the results on AG News and Yelp datasets on fine-tuned BERT and ALBERT model. Our method performs better on both datasets by increasing the success rate by about 1-3\% than the previous methods and also decreasing the perturbation percentage.
Table \ref{tab:mnli} gives the results of attacking a fine-tuned BERT on MNLI.  Although we did not perform better than original BAE-R, we were still able to outperform TextFooler. Due to the unavailability of MNLI fine-tuned ALBERT model in TextAttack, we did not perform an attack on ALBERT. It can also be seen from Table \ref{tab:bertandalbert} that the perturbation percentage for AG's News exceed more than 20\%, which seems to be a perceptible change, but since the average length of the article is only 40.41, making the space for finding relevant words less, the perturbation percentage becomes very high.

To compare attacks across different transformer-based models, we evaluate the performance of Adv-OLM on IMDB dataset. Table \ref{tab:transformersimdb} provides the results of different attack methods on BERT, ALBERT, RoBERTa, DistilBERT and BiLSTM.
Adv-OLM was able to outperform previous attack methods on BERT, ALBERT, RoBERTa by increasing the success rate up to 10\% for BAE-R and up to 6\% for TextFooler. Perturbation percentage was also reduced by 1-2\%. On DistilBERT, Adv-OLM showed no change in the success rate, but the perturbation percentage was lowered slightly.
We also performed an attack on a non-transformer based BiLSTM model which did not show any improvements in the success rate. For BAE-R, it even showed a decrease in the success rate for Adv-OLM. One possible reason for this might be that in both OLM and OLM-S word sampling is performed using a transformer-based BERT language model. We also have qualitative results on IMDB dataset (Figure \ref{fig:textfooler_imdb}, \ref{fig:textfooler_olm_imdb}  in Appendix).

Experimentally it was observed that better words were ranked when OLM/OLM-S was used as the Word Ranking strategy (Figure \ref{fig:textfooler_olm_imdb}). 
When comparing with the original methods, Adv-OLM has more number of queries, which is due to the fact that for word rankings, OLM/OLM-S queries the model a  number of times, thus increasing the overall queries. However, the difference in the number of Adv-OLM queries with the existing attacking methods is not very significant since the model is queried only for unique words from the samples generated from the language model.
\section{Conclusion}
\noindent In this work, we present \textbf{Adv-OLM}, a black box attacking method that uses OLM based word ranking strategy, improving the attack performance significantly over previous methods. We also studied how replacing a single variable in a complex system with a new existing method can improve upon the previously existing attack strategies.
For future work, we would like to experiment with other language models in the OLM algorithm. We plan to study the effect of using different transformers for the language model and the target model. 

\bibliography{anthology,eacl2021}
\bibliographystyle{acl_natbib}

\clearpage
\appendix
\addcontentsline{toc}{section}{Appendix}
\section*{Appendix}
\label{section:appendix}
\section{Datasets}
\label{section:dataset}
We evaluate our adversarial attacks on text classification and natural language inference datasets. We evaluate our method on 500 samples randomly selected from the test set of the given dataset.\\
\textbf{Text classification}  We used the following text classification datasets: 

\begin{itemize}
    \item \textbf{IMDB:} Document-level large Movie Review dataset for binary sentiment classification. \footnote{\href{http://ai.stanford.edu/~amaas/data/sentiment/}{IMDB dataset}}
    \item \textbf{Yelp:} The Yelp reviews dataset consists of reviews from Yelp. This is a dataset for sentiment classfication. It is extracted from the Yelp Dataset Challenge 2015 data. \footnote{\href{http://www.yelp.com/dataset_challenge}{Yelp dataset}}
    \item \textbf{AG's News:} Sentence level news-type classi- fication dataset, containing 4 types of news: World, Sports, Business, and Science.
    \footnote{\href{http://www.di.unipi.it/~gulli/AG_corpus_of_news_articles.html}{AG's News dataset}}
\end{itemize}

\textbf{Natural Language Inference} 
\begin{itemize}
    \item \textbf{MNLI:} The corpus of sentence pairs manually labeled for classification with the labels entailment, contradiction, and neutral, supporting the task of natural language inference (NLI). Unlike SNLI, MNLI is more diverse, based on multi-genre texts, covering transcribed speech, popular fiction, and government reports. \footnote{\href{http://www.nyu.edu/projects/bowman/multinli/}{MNLI dataset}}
\end{itemize}


Average Length is the average number of words in the randomly chosen 500 samples taken from its test set for each dataset.

\section{Textattack}
\label{section:textattack}
TextAttack is an open-source python framework for adversarial attacks, data augmentation and adversarial training in NLP.

\begin{table}[t]\small
\renewcommand{\arraystretch}{1.8}
\centering
\setlength\tabcolsep{3.5pt}
\resizebox{\columnwidth}{!}{
\begin{tabular}{c|c}
\hline
\multicolumn{2}{c}{\textbf{Attack Recipes}}                                         \\ \hline \hline
BAE \citep{garg2020bae} & PWWS \citep{ren2019generating}\\ \hline
Bert-Attack \citep{li2020bert}  & TextFooler \citep{jin2019bert}\\ \hline
DeepwordBug \citep{gao2018black} & HotFlip \citep{ebrahimi2017hotflip} \\ \hline
Alzantot \citep{alzantot2018generating} & Morpheus \citep{tan-etal-2020-morphin} \\ \hline
IGA \citep{wang2019natural} & Pruthi \citep{pruthi2019combating} \\ \hline
Input-Reduction \citep{feng2018pathologies} & PSO \citep{zang2020word}  \\ \hline
Seq2Sick \citep{cheng2020seq2sick} & TextBugger \citep{li2018textbugger}    \\ \hline
Kuleshov \citep{kuleshov2018adversarial} & Fast Alzantot \citep{jia2019certified} \\ \hline
\end{tabular}
}
\caption{Adversarial attacks implemented in Textattack}
\label{table:recipes}
\end{table}

Because of the modularity that TextAttack provides, it enables researchers to construct new attacks from a combination of novel and existing approaches or perform analysis on the already existing approaches. This helps in composing and comparing the attacks in a shared environment. TextAttack makes it easy to perform benchmark comparisons across all the previous attacks performed across models. Text Attack provides clean, readable implementations of 16 adversarial attacks from the literature. Out of which two are sequence to sequence attacks and nine are classification based attacks from the GLUE benchmark. A list of these attacks is presented in Table \ref{table:recipes}. TextAttack is directly integrated with HuggingFace’s transformers and NLP libraries. This allows users to test attacks on models and datasets. 

TextAttack builds attacks from four components:
\begin{enumerate}
   \item A \textbf{search method} that selects the words to be transformed.
    \item A \textbf{transformation} that generates a set of possible perturbations for the given input.
    \item A \textbf{set of constraints} implied on the transformation to ensure that the perturbations are valid with respect to the original input.
    \item A \textbf{goal function} that determines whether an attack is successful in terms of model outputs. For classification tasks, untargeted, and targeted. For a sequence to sequence tasks, non-overlapping output, and minimum BLEU score.
\end{enumerate}

For our approach, we attack TextAttack's fine-tuned models on datasets discussed in \ref{section:dataset}, that are publically available on huggingface\footnote{\href{https://huggingface.co/textattack/}{TextAttack fine-tuned models on HuggingFace}} 
Textattack is also used for the execution of all the previous attacking methods and our Adv-OLM as well.

\section{Word Replacement Strategies}
\label{section:wordreplacement}
\subsection{TextFooler Word Replacement Strategy}
Following workflow was proposed by the paper:

\textbf{Synonym Extraction:} Gather a candidate set CANDIDATES for all possible replacements of the selected word $w_i$ and every other word in the vocabulary.
To represent the words, counter fitting word embeddings were used. Using this set of embedding vectors, top $N$ synonyms whose cosine similarity with $w$ is higher than some $\delta$ were chosen.

\textbf{POS Checking:} In the set CANDIDATES of the word $w_i$, only the ones with the same part-of-speech(POS) as $w_i$ were kept. This step assures that the grammar of the text is mostly maintained.

\textbf{Semantic Similarity Checking:} For each remaining word $c \in $ CANDIDATES, these were substituted for $w_i$ in the sentence $X$, and an adversarial example $X_{adv}$ was obtained. Universal Sentence Encoder (USE) was used to encode the two sentences into high dimensional vectors and then use their cosine similarity score to calculate the sentence similarity between $X$ and $X_{adv}$. The words resulting in similarity scores above a preset threshold $\epsilon$ were placed in a final candidate pool (FINCANDIDATE).

Finally, every candidate word from the FINCANDIDATE was chosen one by one, and the one that 
resulted in the least confidence score of label $y$ was considered as the best replacement for word $w_i$.











\subsection{BAE-R Word Replacement Strategy}
BAE uses a pre-trained BERT masked language model(MLM) to predict the mask tokens for replacement. Since BERT is powerful and trained on the large training corpus, the predicted mask tokens fit well grammatically in the sentence. BERT-MLM does not, however, guarantee semantic coherence to the original text. To ensure semantic similarity on introducing perturbations in the input text, a set of K masked tokens were filtered out using Universal Sentence Encoder(USE) based on sentence similarity score. An additional check for grammatical correctness of the generated adversarial example by filtering out predicted tokens that do not form the same part of speech(POS) as the original token in the sentence was performed.


\section{More Examples}
\label{section:examples}
\begin{figure*}[h!]
\centering
\begin{subfigure}{\textwidth}
  \includegraphics[width=\textwidth]{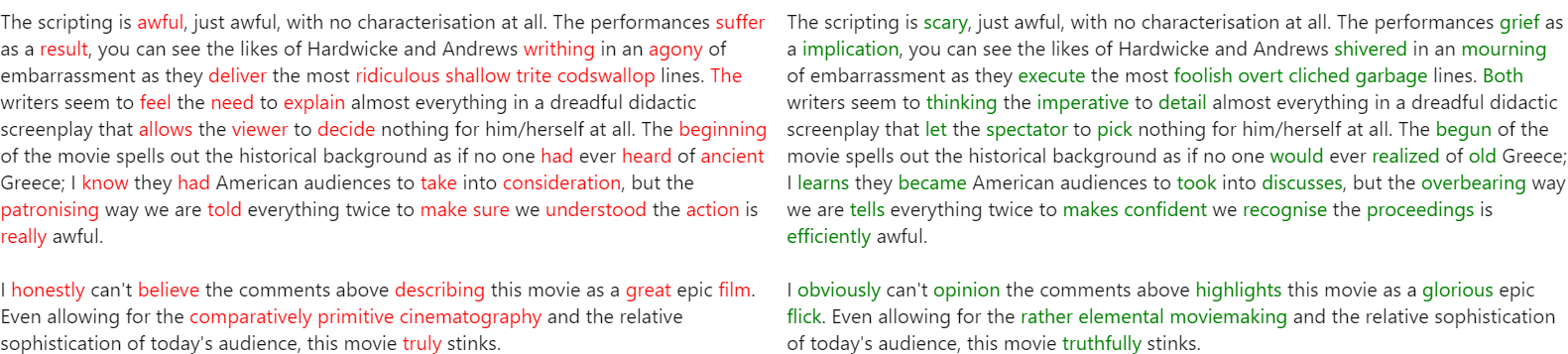}
  \caption{TextFooler Attack on fine-tuned ALBERT on IMDB data sample. $[\textcolor{red}{Negative (100\%)} \rightarrow \textcolor{green}{Positive (51\%)}]$
  }
  \label{fig:textfooler_imdb_albert}
\end{subfigure}

\begin{subfigure}{\textwidth}
  \includegraphics[width=\textwidth]{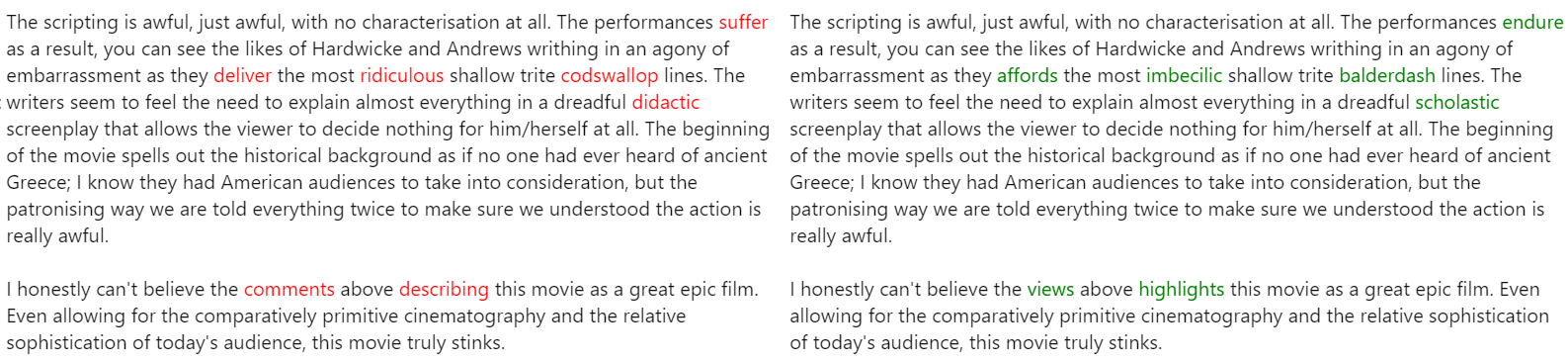}
  \caption{Adv-OLM attack on fine-tuned ALBERT on IMDB data sample.
    $[\textcolor{red}{Negative (100\%)} \rightarrow \textcolor{green}{Positive(57\%)}]$ 
    }
    \label{fig:textfooler_olm_imdb_albert}
\end{subfigure}
\caption{Qualitative Examples of TextFooler and Adv-OLM on ALBERT classifier (\textcolor{red}{Red} words are replaced by \textcolor{green}{Green} words while changing the output prediction probability.)}
\label{fig:textfooler_olm_albert_example}
\end{figure*}

\begin{figure*}[h!]
\centering
\begin{subfigure}{\textwidth}
  \includegraphics[width=\textwidth]{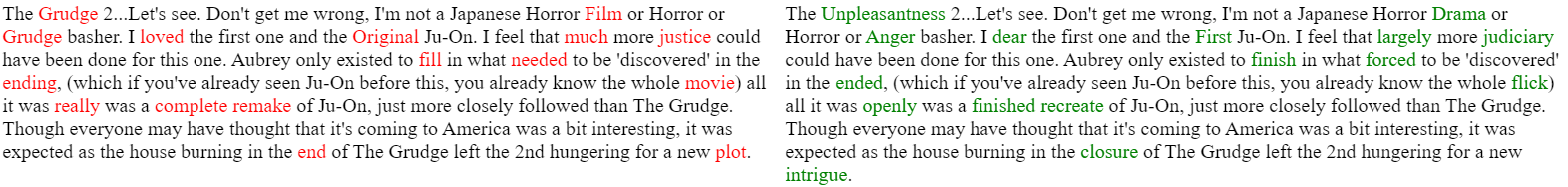}
  \caption{TextFooler Attack on fine-tuned ALBERT on IMDB data sample. $[\textcolor{red}{Negative (100\%)} \rightarrow \textcolor{green}{Positive (51\%)}]$
  }
  \label{fig:textfooler_imdb_albert}
\end{subfigure}

\begin{subfigure}{\textwidth}
  \includegraphics[width=\textwidth]{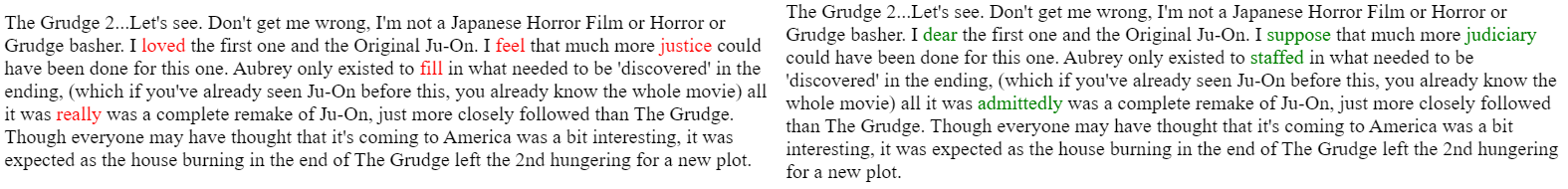}
  \caption{Adv-OLM attack on fine-tuned ALBERT on IMDB data sample.
    $[\textcolor{red}{Negative (100\%)} \rightarrow \textcolor{green}{Positive(65\%)}]$ 
    }
    \label{fig:textfooler_olm_imdb_albert}
\end{subfigure}
\caption{Qualitative Examples of TextFooler and Adv-OLM on RoBERTa classifier (\textcolor{red}{Red} words are replaced by \textcolor{green}{Green} words while changing the output prediction probability.)}
\label{fig:textfooler_olm_albert_example}
\end{figure*}

\end{document}